\documentclass[10pt,a4paper]{article}

\usepackage[final]{cogsci}
\usepackage{amsmath,graphicx,hyperref}
\usepackage{url}            
\usepackage{booktabs}       
\usepackage{amsfonts}       
\usepackage{nicefrac}       
\usepackage{microtype}      
\usepackage{lipsum}
\usepackage{CJKutf8}
\usepackage{amsmath}
\usepackage{multirow}
\usepackage{fancyhdr}       
\graphicspath{{media/}}     
\usepackage{xcolor} 
\usepackage[normalem]{ulem}
\useunder{\uline}{\ul}{}
\usepackage[inline]{enumitem}
\usepackage{enumitem}
\usepackage{graphicx}
\usepackage{boldline}
\usepackage{hhline}

\usepackage[
  style=apa,
  natbib=true,
  annotation=false,
]{biblatex}
\usepackage{float} 

\title{From Prediction to Justification: Aligning Sentiment Reasoning with Human Rationale via Reinforcement Learning}

\author{
  {\large\bfseries Shihao Zhang$^1$, Ziwei Wang$^1$, Jie Zhou$^1$\thanks{Corresponding authors, jzhou@cs.ecnu.edu.cn, lyyu@cs.ecnu.edu.cn.}, Yulan Wu$^1$, Qin Chen$^1$,} \\ {\large\bfseries Zhikai Lei$^2$, Liyang Yu$^{1,3*}$, Liang Dou$^2$, Liang He$^2$} \\
  {\normalsize\normalfont
    $^1$School of Computer Science and Technology, East China Normal University, \\ $^2$Shanghai Qiji Zhifeng Co., Ltd. $^3$Ocean University of China
  }
}

\begin{document}

\maketitle

\begin{abstract}

While Aspect-based Sentiment Analysis (ABSA) systems have achieved high accuracy in identifying sentiment polarities, they often operate as "black boxes," lacking the explicit reasoning capabilities characteristic of human affective cognition. Humans do not merely categorize sentiment; they construct causal explanations for their judgments. To bridge this gap, we propose ABSA-R1, a large language model framework designed to mimic this ``reason-before-predict" cognitive process. By leveraging reinforcement learning (RL), ABSA-R1 learns to articulate the why behind the what, generating natural language justifications that ground its sentiment predictions. We introduce a Cognition-Aligned Reward Model (formerly sentiment-aware reward model) that enforces consistency between the generated reasoning path and the final emotional label. Furthermore, inspired by metacognitive monitoring, we implement a performance-driven rejection sampling strategy that selectively targets hard cases where the model's internal reasoning is uncertain or inconsistent. Experimental results on four benchmarks demonstrate that equipping models with this explicit reasoning capability not only enhances interpretability but also yields superior performance in sentiment classification and triplet extraction compared to non-reasoning baselines.

\end{abstract}

\section{Introduction}
Aspect-Based Sentiment Analysis (ABSA) is a fine-grained sentiment understanding task that aims to identify the sentiment polarity expressed toward specific aspects in a given text \citep{zhou2019deep}. For example, in the sentence ``The battery life is great, but the screen is dim," ABSA systems are expected to recognize that ``battery life" is associated with positive sentiment, while ``screen" elicits negative sentiment. Beyond simple classification, more advanced variants such as Aspect-Opinion-Sentiment Triplet Extraction (AOSTE) \citep{xu2021learning, chen2022enhanced} further require models to jointly extract aspect terms, opinion expressions, and their sentiment polarities, enabling a comprehensive cognitive mapping of subjective language.

\begin{figure}[t!]
  \centering
  \includegraphics[width=0.95\linewidth]{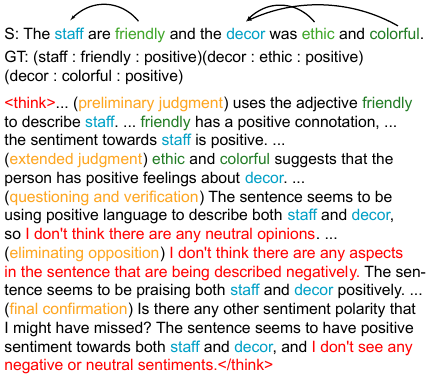}
  \vspace{-2mm}
  \caption{An example of sentiment thinking.}
  \label{fig:intro}
  \vspace{-3mm}
\end{figure}

Over the years, significant progress has been made in ABSA through both traditional supervised learning \citep{pontiki-etal-2014-semeval,mohammad-etal-2013-nrc} and pre-trained language models such as BERT and its domain-adapted variants \citep{DBLP:journals/corr/abs-1810-04805}. These methods typically treat ABSA as a statistical classification or sequence labeling problem, achieving strong performance in label prediction \citep{li2019unified}. Furthermore, large language models (LLMs) such as T5 \citep{chung2022scaling} and LLaMA \citep{metaai2024introducing} have been utilized to generate aspects, opinions, and sentiments via instruction tuning \citep{scaria-etal-2024-instructabsa,10507027}. However, a critical cognitive gap remains: these models often operate as opaque black boxes. Unlike humans, who construct internal justifications before reaching a sentiment judgment, these models output labels without providing human-intelligible rationales. This lack of transparency limits their interpretability, failing to align with the explicability inherent in human affective cognition.


Recently, reasoning LLMs equipped with chain-of-thought or reinforcement learning have achieved impressive performance in complex tasks across domains. These models do not merely predict outcomes; they articulate the logical steps leading to them, mirroring the ``System 2'' thinking process in human cognition. This progress motivates a critical question at the intersection of NLP and cognitive modeling: \textbf{Can we endow LLMs with the ability to explicitly learn the affective reasoning processes humans use to attribute sentiment?} For example, as shown in Fig. \ref{fig:intro}, the model should provide a reasoning path to explain \textit{why} a user feels positively about staff and decor, rather than just identifying \textit{what} the sentiment is.

In this work, we propose ABSA-R1, a novel sentiment reasoning LLM for ABSA. Trained via Reinforcement Learning (RL), our model adopts a ``reason before prediction" cognitive paradigm, learning to generate natural language explanations that serve as the causal basis for its sentiment predictions. To guide this process, we design a cognition-aligned reward model (formerly sentiment-aware reward model) that jointly assesses the factual accuracy of the prediction and the logical coherence of the explanation, ensuring alignment between the generated rationale and the emotional label. Furthermore, we introduce a performance-driven rejection sampling strategy that functions as a metacognitive filter. This strategy dynamically identifies ``hard'' training samples—instances where the model's internal reasoning is ambiguous or inconsistent—enabling more effective learning from cognitively challenging cases.

Our main contributions are threefold: 
\begin{itemize}[leftmargin=*, align=left]
    \item We propose ABSA-R1, an RL-based LLM framework explicitly designed to mimic human-like sentiment reasoning in ABSA, generating interpretable justifications that ground its predictions.
    \item We introduce a cognition-aligned reward model that ensures the synchronization of sentiment prediction with high-quality reasoning, alongside a performance-driven rejection sampling strategy to enhance robustness on ambiguous samples.
    \item Extensive experiments on benchmark datasets demonstrate that ABSA-R1 achieves SOTA performance in both ABSC and AOSTE, while providing transparent, human-readable reasoning traces that bridge the gap between machine prediction and human understanding.
\end{itemize}

\section{Related Work}
\subsection{Aspect-Based Sentiment Analysis} 
Aspect-Based Sentiment Analysis (ABSA), a core task in fine-grained sentiment understanding, focuses on parsing the sentiment orientations towards different aspects in text \citep{liu2024let}. 
This task has evolved through several stages: from early rule- and lexicon-based methods with limited generalization \citep{hu2004mining}, to machine learning approaches leveraging SVMs and CRFs with handcrafted features \citep{dong-etal-2014-adaptive,wang2016attention}, and more recently, deep learning models employing RNNs, attention mechanisms, and graph neural networks to capture contextual and syntactic dependencies \citep{tang2016effective,xue2018aspect}. 
With the rise of pre-trained language models (PLMs), fine-tuning BERT-based architectures has become a dominant paradigm, significantly boosting performance across ABSA subtasks including aspect extraction, aspect sentiment classification (ASC), and aspect-opinion sentiment triplet extraction (ASTE) \citep{xu2021learning,chen2022enhanced,mao2021joint}. More recent work further casts ABSA as a sequence-to-sequence generation problem using models like T5 or large language models (LLMs) such as ChatGPT and LLaMA, enabling unified multi-task frameworks \citep{shen-etal-2021-enhancing,scaria-etal-2024-instructabsa}. Despite these advances, most existing methods remain focused on prediction accuracy and lack mechanisms for generating interpretable reasoning, offering no insight into why a particular sentiment is assigned, which limits their transparency and trustworthiness in real-world applications.

\subsection{Reasoning LLMs}

Recent advances in LLMs have demonstrated that explicit reasoning, particularly through Chain-of-Thought prompting, can significantly enhance performance on complex reasoning tasks by encouraging models to generate intermediate justifications before producing final answers \citep{wei2023chainofthoughtpromptingelicitsreasoning,kojima2023largelanguagemodelszeroshot}. To train models for reasoning rather than mere pattern matching, reinforcement learning frameworks such as RLHF \citep{10.5555/3600270.3602281} and its variants (e.g., PPO \citep{schulman2017proximalpolicyoptimizationalgorithms}, DPO \citep{rafailov2024directpreferenceoptimizationlanguage}) have been widely adopted. These methods optimize models based on learned or engineered reward signals that reflect human preferences or task-specific correctness. More recently, rule-based or logic-aware reward models have emerged to provide more precise and interpretable feedback, especially in structured reasoning domains \citep{yu2025dapoopensourcellmreinforcement,NEURIPS2024_c4e380fb,hu2025openreasonerzeroopensourceapproach}. Techniques like STaR \citep{NEURIPS2022_639a9a17} and GRPO \citep{shao2024deepseekmathpushinglimitsmathematical} further enable models to iteratively improve reasoning through self-training and reward-guided policy updates. While these approaches have shown promise in mathematical and commonsense reasoning, their application to sentiment analysis, particularly in generating meaningful, sentiment-grounded explanations, remains underexplored.

\begin{figure*}[t!]
  \centering
  \includegraphics[width=0.94\linewidth]{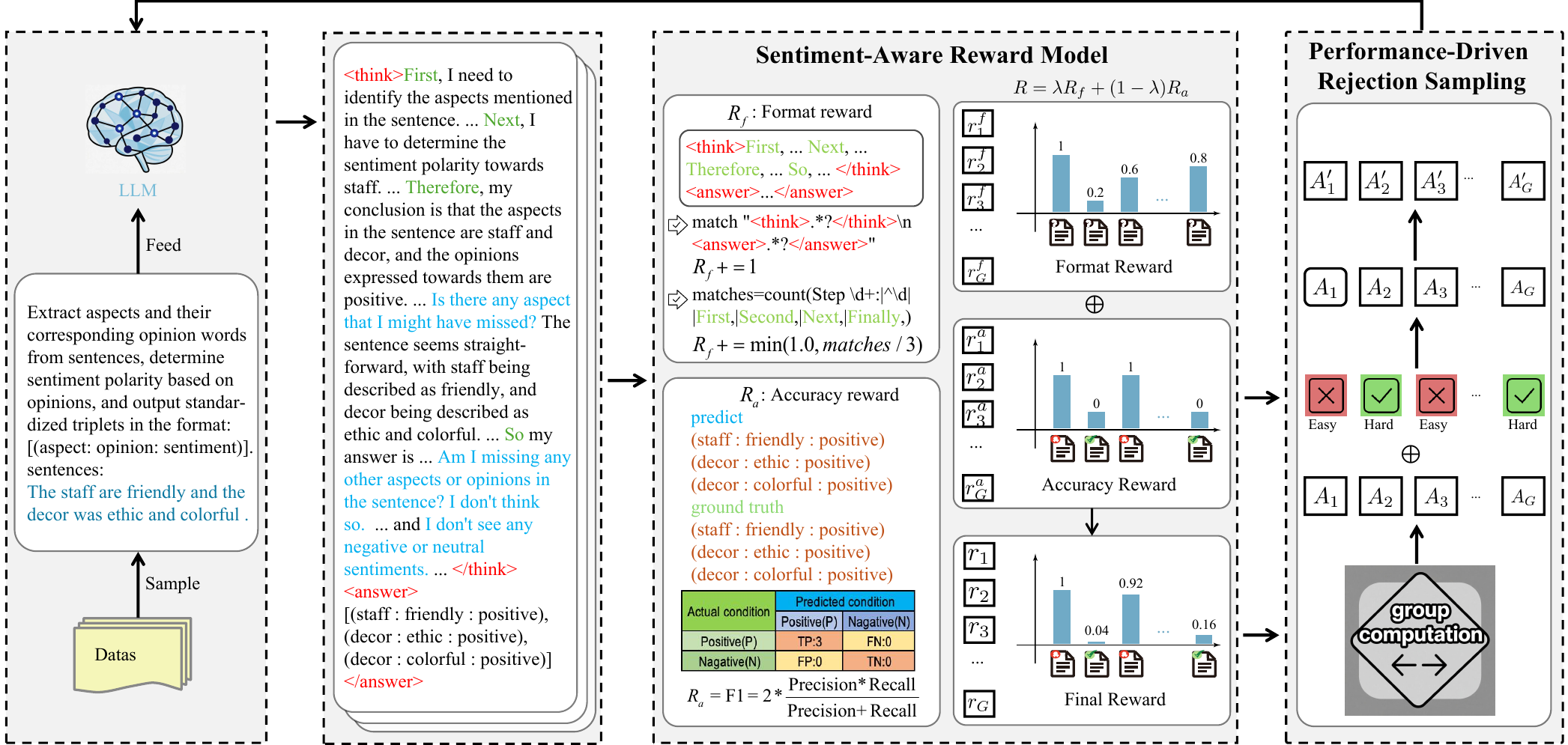}
  \vspace{-1.5mm}
  \caption{The framework of our ABSA-R1, which simulates a System 2 thinking process by generating explicit rationales before making predictions. A Sentiment-Aware Reward Model provides feedback on the structural coherence of the reasoning and the correctness of the sentiment. To enhance robustness, the Performance-Driven Rejection Sampling acts as a metacognitive filter, selecting challenging instances where the model's reasoning is inconsistent for targeted reinforcement learning.}
  \label{fig:framework}
  \vspace{-2mm}
\end{figure*}

\section{Our Method}
\label{sec:method}
We propose ABSA-R1, a novel reinforcement learning (RL)-based framework for Aspect-Based Sentiment Analysis that explicitly aligns sentiment prediction with generated reasoning traces (see Fig.~\ref{fig:framework}).
Operating under a \emph{reasoning-before-prediction} cognitive paradigm, our model transforms the traditional direct-mapping approach into a structured generative process.
Specifically, given an input $x$, which can be a sentence with a target aspect (for ABSC) or raw text requiring element extraction (for AOSTE), ABSA-R1 generates a composite output $o = (p, \hat{y})$.
Here, $p$ represents a natural language reasoning path that articulates the causal logic behind the sentiment judgment, and $\hat{y}$ denotes the final decision, which manifests as either a sentiment polarity label or a set of extracted sentiment triplets.
By enforcing the generation of $p$ prior to $\hat{y}$, we ensure that the model's predictions are grounded in explicit intermediate reasoning, thereby enhancing both the transparency of the decision-making process and the interpretability of the results for human end-users.

The model $\pi_\theta$ parameterized by $\theta$ is trained using an improved variant of Generalized Reward Policy Optimization \citep{shao2024deepseekmathpushinglimitsmathematical} to mitigate common RL issues such as premature overfitting and entropy collapse, which degrade generalization in reasoning tasks. 
The overall training objective is:
\begin{equation}
\small
\nonumber
\begin{aligned}
\mathcal{L}(\theta) =\quad& \mathbb{E}_{(x)\sim \mathcal{D}, \{O_i\}_{i=1}^G\sim \pi_{\theta_\text{old}}(\cdot\mid x)}
\Bigg[\frac{1}{\sum_{i=1}^{G}|O_i|}\sum_{i=1}^{G}\sum_{t=1}^{|O_i|} 
\\&  \min \Big( r_{i,t}(\theta) \hat{A}_{i,t},  
\ \text{clip} \Big( r_{i,t}(\theta), 1 - {\varepsilon_{\text{low}}}, 1 + {\varepsilon_{\text{high}}} \Big) \hat{A}_{i,t} \Big)\Bigg]
\end{aligned}
\label{eq:GRPOloss}
\end{equation}
where the probability ratio is defined as:
\begin{equation}
    r_{i,t}(\theta) = \frac{\pi_{\theta}(O_{i,t} \mid x, O_{i,<t})}{\pi_{\theta_{\text{old}}}(O_{i,t} \mid x, O_{i,<t})},
\label{eq:ratio}
\end{equation}
and the advantage $\hat{A}_{i,t}$ is normalized across $G$ generated reasoning paths:
\begin{equation}
    \hat{A}_{i,t} = \frac{R_i - \text{mean}(\{R_i\}_{i=1}^G)}{\text{std}(\{R_i\}_{i=1}^G) + \epsilon}.
\label{eq:advantage_calculation}
\end{equation}

In the following, we present our sentiment-aware reward model and performance-driven rejection sampling strategy, which together guide the model to generate both accurate and well-structured reasoning.

\subsection{Sentiment-aware Reward Model}
\label{subsec:reward}

To ensure high-quality reasoning and avoid reward hijacking, where models exploit shallow patterns to maximize reward without real understanding, we design a rule-based, customizable reward function tailored to the structured nature of ABSA tasks. Unlike black-box reward models, our approach provides explicit, interpretable feedback grounded in task semantics.

The reward for a generated output $o$ (including reasoning $p$ and prediction $\hat{y}$) given ground truth $y$ is defined as:
\begin{equation}
    R(o, y) = \lambda \cdot R_f(p) + (1-\lambda) \cdot R_a(\hat{y}, y),
\end{equation}
where $\lambda$ balances the importance of format and correctness.

\textbf{Format Reward $R_f$.}
This component ensures the model adheres to a structured reasoning format. It consists of three sub-rewards: 1) Tag Compliance: The output must include designated placeholders such as \texttt{$<$think$>$} and \texttt{$<$answer$>$} to separate reasoning from prediction; 2) Logical Flow: Use of transitional phrases (e.g., “firstly”, “therefore”) is encouraged to promote coherent, step-by-step reasoning; 3) Structural Integrity: The number and order of tags are validated; deviations incur penalties.
The format reward is computed as a weighted sum of these sub-components, normalized to $[0,1]$.

\textbf{Answer Reward $R_a$.}
This evaluates the correctness of the final sentiment prediction or extracted sentiment triplets.

For ABSC, we use exact match accuracy:
\begin{equation}
    R_a(\hat{y}, y) = 
    \begin{cases} 
    1, & \texttt{is\_equivalent}(\hat{y}, y) \\
    0, & \text{otherwise}
    \end{cases}
\end{equation}

For AOSTE, we compute a soft F1-based reward. A predicted triplet $\hat{y}$ is considered a true positive (TP) if both aspect and opinion are substrings of the corresponding ground-truth terms:
\begin{equation}
    \text{TP} = 
    \begin{cases} 
    1, & \texttt{is\_substrings}(\hat{y}, y) \\
    0, & \text{otherwise}
    \end{cases}
\end{equation}
Precision, recall, and F1 are computed over all predicted and ground-truth triplets. To balance precision and recall and penalize excessive hallucination or omission, we introduce a soft penalty:
\begin{equation}
    R_a(\hat{y}, y) = \text{F1} - \gamma \cdot |FN - FP|,
\end{equation}
where $\gamma$ is a small constant (e.g., 0.05) to avoid over-penalization. This ensures the model is rewarded not only for correctness but also for completeness and conciseness.

\subsection{Performance-Driven Rejection Sampling}
\label{subsec:rejection}

To direct learning toward meaningful mistakes and avoid reinforcing already-correct predictions, we introduce a performance-driven rejection sampling strategy that operates at the \emph{generation level}. For each input $x$, instead of treating all generated trajectories equally, we selectively retain only those that are \emph{incorrect} or \emph{incomplete}, while rejecting generations that achieve full correctness.

Concretely, given an input $x$, we sample $G$ candidate outputs $\{O_i\}_{i=1}^G$ from the current policy $\pi_\theta(\cdot \mid x)$. Each $O_i$ includes both a reasoning trace and a final prediction. We then evaluate whether $O_i$ is correct using a task-specific metric:
\[
c_i = \texttt{is\_correct}(O_i, y),
\]
where $\texttt{is\_correct}(\cdot)$ returns 1 if the prediction matches the gold label (exact match for ABSC, F1-based threshold for AOSTE), and 0 otherwise.

We filter the generated samples as follows: If $c_i = 1$, i.e., the $i$-th generation is fully correct, we \emph{discard} this trajectory.
If $c_i = 0$, i.e., the reasoning or prediction is flawed, we \emph{retain} it for reinforcement learning. The final training batch consists only of incorrect generations $\{O_i \mid c_i = 0\}$, ensuring that the policy update is driven exclusively by the model’s mistakes.
By learning from its own errors, ABSA-R1 effectively turns failed generations into valuable learning signals, accelerating convergence toward robust and accurate reasoning.
Moreover, since incorrect generations often reflect deeper semantic or structural misunderstandings, training on them enhances the model’s ability to handle challenging cases, including implicit sentiment, negation, and compositional reasoning.

\section{Experimental Setups}
\subsection{Dataset}
Following prior work~\citep{DBLP:journals/corr/abs-2412-00763,scaria-etal-2024-instructabsa}, we evaluate our model on the SemEval benchmarks
~\citep{pontiki-etal-2014-semeval,pontiki-etal-2015-semeval,pontiki-etal-2016-semeval}, which are widely used in ABSA. 
To facilitate reasoning-aware training, 
we sample 10\% of the original training set to construct a cold-start dataset, where reasoning chains are automatically generated using DeepSeek-R1 \citep{deepseekai2025deepseekr1incentivizingreasoningcapability}. 

\subsection{Baselines}
We compare ABSA-R1 against a comprehensive set of baselines across two main tasks.

For Aspect-Opinion-Sentiment Triplet Extraction (AOSTE), we include structured models such as:
\begin{itemize}[leftmargin=*, align=left]
    \item \textbf{BARTABSA}~\citep{yan-etal-2021-unified}: A unified generative framework that converts ABSA tasks into a sequence-to-sequence generation problem using BART to produce sentiment triplets as a structured sequence.
    \item \textbf{LEGO-ABSA}~\citep{gao-etal-2022-lego}: A prompt-based, task-assemblable generative framework that treats different ABSA subtasks as "Lego" blocks, enabling multi-task learning and transfer from simple to complex tasks.
    \item \textbf{Span-ASTE}~\citep{xu-etal-2021-learning}: A span-level model that captures target-opinion span interactions via ATE/OTE-supervised dual-channel pruning for end-to-end extraction of aspect sentiment triplets.
    \item \textbf{SBN} (Span-level Bidirectional Network)~\citep{chen2022span}: A model that utilizes span-level representations to extract triplets by decoding from both aspect-to-opinion and opinion-to-aspect directions.
    \item \textbf{ParaPhrase-T5}~\citep{zhang2021aspect}: A generative approach that reformulates the Aspect Sentiment Quad Prediction (ASQP) task as a paraphrase generation problem using T5.
    \item \textbf{TAGS}~\citep{xianlong-etal-2023-tagging}: A tagging-assisted generation model that incorporates encoder and decoder supervision to guide the generation of sentiment triplets.
    \item \textbf{EHG} (Efficient Hybrid Generation)~\citep{lv-etal-2023-efficient}: A hybrid framework combining efficient generation mechanisms with structured prediction to enhance performance on ABSA tasks.
    \item \textbf{MvP} (Multi-view Prompting)~\citep{gou2023mvp}: A method that leverages multi-view prompting to aggregate sentiment elements generated in varying orders, improving robustness and accuracy in tuple prediction.
    \item \textbf{PGSO}~\citep{DBLP:journals/corr/abs-2412-00763}: A prompt-based generative sequence optimization network that dynamically regulates the generation process to better align with aspect-based sentiment structures.
\end{itemize}

We also evaluate state-of-the-art LLMs, including T5-Instruct~\citep{chung2022scaling}, ChatGLM3-6B~\citep{du-etal-2022-glm}, Mistral-7B-v0.2~\citep{jiang2023mistral}, LLaMA3-8B~\citep{metaai2024introducing}, and Qwen2.5-7B~\citep{qwen2.5}, acting as general-purpose baselines.

For Aspect-Based Sentiment Classification (ABSC), we compare with classical neural models, including:
\begin{itemize}[leftmargin=*, align=left]
    \item \textbf{IAN} (Interactive Attention Networks)~\citep{10.5555/3171837.3171854}: A model employing two attention networks to interactively learn representations of the aspect and the context.
    \item \textbf{PGCNN}~\citep{huang-carley-2018-parameterized}: A CNN with a parametric gating unit that extracts aspect term features for effective aspect-level sentiment analysis.
    \item \textbf{CPA-SA}~\citep{10.1016/j.knosys.2022.108473}: An approach focusing on aspect-specific context position information to enhance sentiment classification accuracy.
    \item \textbf{RMN-P}~\citep{zeng-etal-2022-relation}: A model that uses aspect relation construction as an auxiliary task and bi-attention to capture bidirectional semantics of context and aspect words.
    \item \textbf{KGAN}~\citep{10.1109/TKDE.2023.3250499}: A knowledge graph augmented network that incorporates external knowledge to enrich the semantic representation of aspects and contexts.
    \item \textbf{MGGCN-BERT}~\citep{xiao2022multi}: A method combining multi-head self-attention with gated graph convolutional networks (GCN) to capture syntactic and semantic dependencies.
    \item \textbf{ASHGAT-BERT}~\citep{OUYANG2024123412}: An aspect-specific hypergraph attention network designed to model high-order interactions between words and aspects.
    \item \textbf{CKG-BERT}~\citep{gao2024ckg}: A framework that improves ABSA by leveraging text augmentation via ChatGPT and a knowledge-enhanced gated attention graph convolutional network.
\end{itemize}

As well as recent instruction-tuned approaches, including:
\begin{itemize}[leftmargin=*, align=left]
    \item \textbf{InstructABSA2}~\citep{scaria-etal-2024-instructabsa}: An instruction-learning framework tailored for ABSA that fine-tunes LMs to follow task-specific sentiment instructions.
    \item \textbf{InstructSFT}~\citep{qwen2.5}: A supervised fine-tuning baseline based on the Qwen2.5 architecture, trained on standard ABSA instruction datasets.
\end{itemize}

\subsection{Implementation Details}
We use Qwen2.5-7B-Instruct as the base LLM, trained within the Verl framework\footnote{\url{https://github.com/volcengine/verl}}. The learning rate is set to $1 \times 10^{-6}$. During the rollout phase, the batch size is 512, and we sample 8 responses per sample to ensure diverse reasoning trajectories for RL. 
All experiments are conducted with distributed training on multiple GPUs, and hyperparameters are selected based on validation performance.

\begin{table}[t!]
    \centering
    \setlength{\tabcolsep}{1.0mm}
    \begin{tabular}{lccccc}
        \toprule
        \multirow{1}{*}{Method} & Lap14& Rest14& Rest15 & Rest16 & Avg \\
        \hline
        T5-Instruct & 60.86& 72.82& 64.86&72.57 & 67.78 \\
        ChatGLM3-6B & 62.05& 74.65& 67.11&70.30  & 68.53\\
        Mistral-7B-v0.2 & 63.83& 76.44& 71.07&76.99 & 72.08 \\
        LLaMA3-8B & 65.68& {79.20} & 74.06&76.15  & 73.77 \\
        Qwen2.5-7B & {68.06}& 79.60& 74.55& 80.91  & 75.78\\
        \hline
        BARTABSA & 57.59& 72.46& 60.11&69.98  & 65.04 \\
        LEGO-ABSA & 62.20& 73.70& 64.40&69.90  & 67.55 \\
        Span-ASTE & 59.38& 71.85& 63.27&70.26 & 66.19\\
        SBN & 62.65& 74.34& 64.82&72.08  & 68.47\\
        ParaPhrase-T5 & 61.13& 72.03& 62.56&71.70  & 66.86\\
        TAGS & 64.53& 75.05& 67.90&76.61  & 71.02\\
        EHG & 61.53& 71.82& 63.58&72.35  & 67.32\\
        MvP & 63.33& 74.05& 65.89&73.48  &69.19 \\
        PGSO& 64.14& 74.38& 67.28&75.33  & 70.28 \\  
        \hline
        ABSA-R1 & \textbf{69.37}& \textbf{81.47}& \textbf{84.81}& \textbf{84.52}  & \textbf{80.04}\\
        ~~w/o PRS & 57.17&74.36 & 73.90&71.10  & 69.13\\
        ~~w/o SAR & 58.80& 72.22& 74.15& 76.45  & 70.40\\
         \bottomrule
    \end{tabular}
    \vspace{-1mm}
    \caption{Experimental results of AOSTE in terms of F1.}
    \label{tab:results_AOSTE}
    \vspace{-3mm}
\end{table}

\section{Experimental Results}
\subsection{Main Results} 

From the experimental results shown in Tables~\ref{tab:results_AOSTE} and \ref{tab:results_ABSC}, our proposed ABSA-R1 demonstrates strong performance across both AOSTE and ABSC tasks.

First, in the AOSTE task (Table~\ref{tab:results_AOSTE}), ABSA-R1 establishes a new state-of-the-art by significantly outperforming both structured models and general-purpose LLMs. 
Specifically, it achieves an average F1 score of 80.04, surpassing the strongest structured baseline (TAGS, 71.02) by 9.02 points and the best LLM baseline (Qwen2.5-7B, 75.78) by 4.26 points. 
The advantage is most pronounced on the Rest15 benchmark, where ABSA-R1 reaches 84.81 F1, exceeding Qwen2.5-7B by 10.26 points. 
These results indicate that our RL-based reasoning framework is particularly effective in handling complex extraction tasks that require deep semantic understanding and structural alignment.

Second, regarding the ABSC task (Table~\ref{tab:results_ABSC}), ABSA-R1 achieves superior performance despite the task's simpler classification nature. 
Our model secures the highest Accuracy on three out of four datasets (Rest14, Rest15, Rest16) and attains the best overall performance with an average Accuracy of 89.95 and F1 of 80.88. 
It is worth noting that ABSA-R1 outperforms the instruction-tuned baseline, InstructSFT, by a large margin ($+5.59$ in Avg F1), demonstrating that incorporating reasoning traces significantly benefits classification accuracy compared to direct answer generation. 
Even when compared to highly specialized architectures like CKG-BERT, ABSA-R1 maintains a lead in average metrics, showcasing its robustness and generalizability without relying on external knowledge bases or complex graph structures.

\begin{table*}[t!]
    \centering
    \setlength{\tabcolsep}{3.0mm}
    \begin{tabular}{lcccccccccc}
        \toprule
        \multirow{2}{*}{Method} & \multicolumn{2}{c}{Rest14} & \multicolumn{2}{c}{Lap14} & \multicolumn{2}{c}{Rest15} & \multicolumn{2}{c}{Rest16} & \multicolumn{2}{c}{Avg} \\
        \cline{2-11}
         & Acc & F1 & Acc & F1 & Acc & F1 & Acc & F1 & Acc & F1 \\
         \hline
         IAN& 80.18& 69.78& 71.36& 68.85& 76.54& 55.62& 83.79&58.62 & 77.97  & 63.22 \\
         PGCNN& 78.43& 69.58& 69.72& 67.13& 75.01& 51.83& 81.34&58.23 &  76.12	 & 61.69 \\
         CPA-SA& 82.64& 73.38& 75.18& 71.50 & 79.61& 60.15& 88.92&72.43 &  81.59	 & 69.37 \\
         RMN-P& 81.16& 73.17& 74.50& 69.79& 80.69& 64.41& 88.75&71.54 & 81.28 &	69.73\\
         KGAN& 87.15& 82.05& 82.66& 78.98& 86.21& 74.20 & 92.34&81.31  & 87.09 &	79.14  \\
         MGGCN-BERT& 83.21& 75.38& 79.57& 76.30& 82.90& 69.27& 89.66&73.99 & 83.84	& 73.74\\
        ASHGAT-BERT & 85.49 & 79.23 & 79.98 & 76.58 & 83.57 & 71.15 & 90.75 & 77.57  &  84.95 &	76.13\\
        CKG BERT & \underline{89.24} & \underline{83.31} & \textbf{83.41} & \textbf{79.41} & 88.92 & \underline{76.42} & 92.87 & \textbf{81.95}  & \underline{88.61}	& \underline{80.27} \\
        InstructABSA2 & 85.17 & - & 81.56 & - & 84.50 & - & 89.43 & -  & 85.16 & -\\
        InstructSFT & 86.88 & 77.36 & 82.29 & 77.45 & \underline{89.78} & 73.66 & \underline{93.38} & 72.70  & 88.08	& 75.29 \\
        \hline
        ABSA-R1 & \textbf{91.32} & \textbf{85.73} & \underline{83.07} & \underline{79.36} & \textbf{90.79}& \textbf{76.73}& \textbf{94.62}& \underline{81.69} & \textbf{89.95} &	\textbf{80.88} \\
        ~~w/o PRS &  88.67& 80.93& 80.09& 66.34& 86.60& 62.31& 93.38&73.18 & 87.18	& 70.69\\
         ~~w/o SAR &  90.50& 84.33& 83.86& 81.01& 89.95& 75.34& 93.69&73.58 &  89.50	 & 78.56 \\
         \bottomrule
    \end{tabular}
    \caption{Experimental results of ABSC.}
    \label{tab:results_ABSC}
    \vspace{-1mm}
\end{table*}

\subsection{Ablation Studies}
To verify the contribution of each component, we conducted ablation studies by removing the Performance-driven Rejection Sampling (PRS) and the Sentiment-Aware Reward (SAR) individually. The results on both AOSTE and ABSC tasks (Table \ref{tab:results_AOSTE} and Table \ref{tab:results_ABSC}) confirm that our proposed mechanisms significantly contribute to the model's reasoning capability.

\textbf{Effectiveness of PRS.}
The proposed PRS strategy is designed to force the model to learn from challenging cases. The removal of PRS (\textit{w/o PRS}) leads to a marked performance drop. 
In the AOSTE task, which requires extracting complex triplets, the model without PRS suffers a loss of over 10 points on the Lap14 dataset ($69.37 \rightarrow 57.17$). 
Furthermore, in the ABSC task, the lack of PRS results in catastrophic failure on determining fine-grained sentiment boundaries, evidenced by the average F1 score dropping from $80.88$ to $70.69$. 
These results strongly suggest that standard sampling is insufficient for sentiment reasoning, and our rejection sampling strategy effectively improves learning efficiency by focusing on error-prone instances.

\textbf{Effectiveness of SAR.}
The SAR module ensures that the generated reasoning logically supports the sentiment prediction. 
Comparison with the \textit{w/o SAR} variant shows that performance consistently lags behind the full model. For example, on the Rest14 AOSTE benchmark, the F1 score decreases by $9.25$ points ($81.47 \rightarrow 72.22$). 
Even on the classification-heavy ABSC task, removing SAR degrades accuracy (e.g., Rest14 Accuracy drops from $91.32$ to $90.50$). 
This indicates that the sentiment-aware reward provides a critical supervision signal, guiding the LLM to generate valid justifications rather than hallucinated reasoning paths.

\subsection{Further Analysis}
\paragraph{Case Studies.}
As illustrated in Figure~\ref{fig:intro}, ABSA-R1 demonstrates a sophisticated cognitive trajectory when analyzing the sentence ``The staff are friendly and the decor was ethic and colorful.'' 
Rather than jumping to a conclusion, the model initiates a \textbf{structured reasoning process}. It begins with a \textit{preliminary judgment} linking ``friendly'' to a positive stance on ``staff,'' followed by an \textit{extended judgment} regarding the ``decor.'' 
Crucially, the model exhibits \textbf{metacognitive monitoring} through the phases of \textit{questioning and verification} and \textit{eliminating opposition}. As seen in the trace, it explicitly asks, ``Is there any other sentiment polarity that I might have missed?'' and actively verifies the absence of negative opinions. 
This self-reflective mechanism, checking for counter-evidence before \textit{final confirmation}, mirrors System 2 human thinking, ensuring that the final triplet extraction is not only accurate but logically robust and transparent. 


\begin{figure}[t!]
    \centering
    \begin{minipage}[b]{0.26\textwidth}
        \centering
        \includegraphics[width=\textwidth]{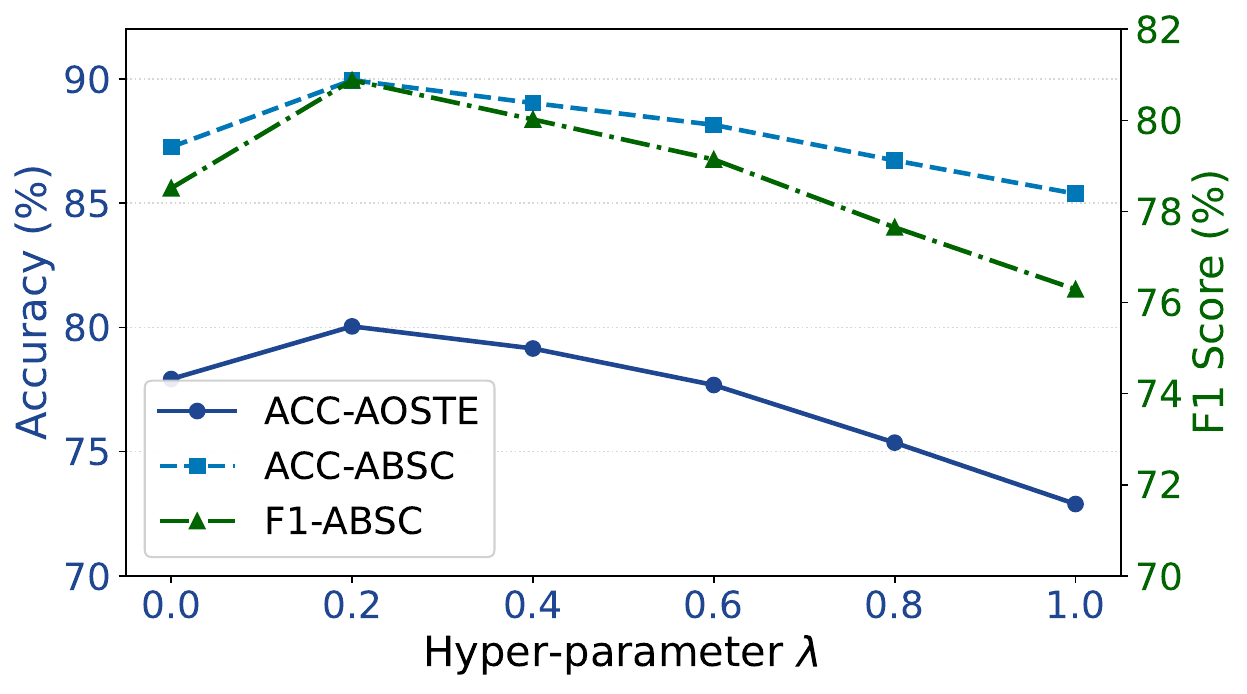}
        \vspace{-6mm}
        \caption{Impact of Hyper-Parameter $\lambda$.}
        \label{fig:hyperparameter}
    \end{minipage}   
    \begin{minipage}[b]{0.21\textwidth}
        \centering
        \includegraphics[width=\textwidth]{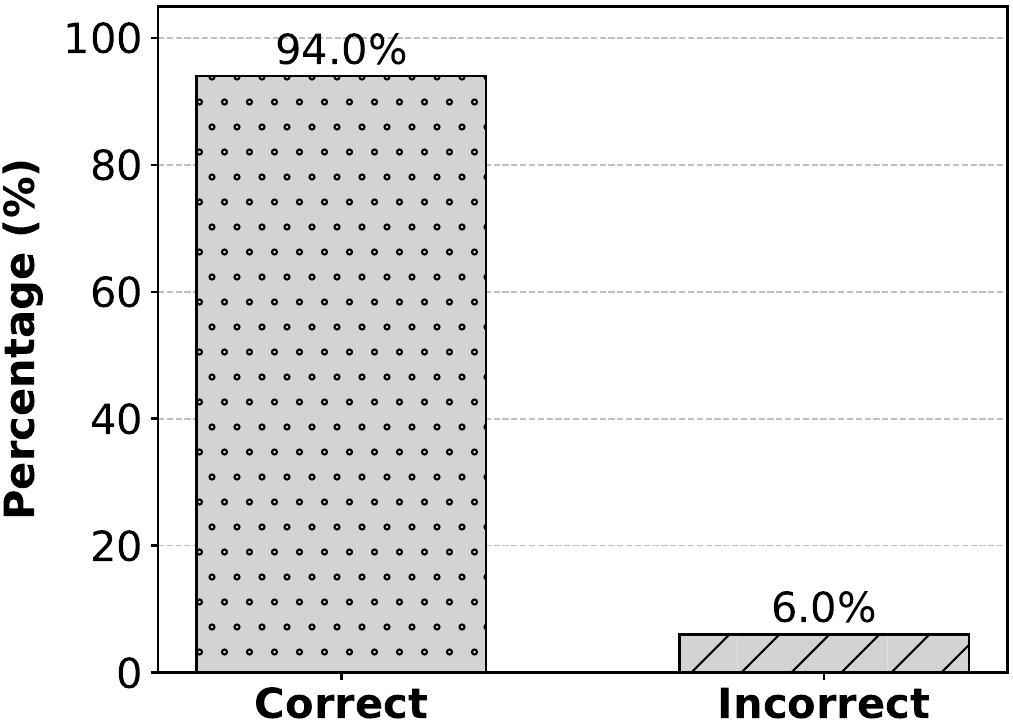}
        \vspace{-6mm}
        \caption{Evaluation of Sentiment Reasoning.}
        \label{fig:human_eval}
    \end{minipage}
    \vspace{-2mm}
\end{figure}

\paragraph{Impact of Hyper-Parameter $\lambda$.}
We analyze the sensitivity of the hyper-parameter $\lambda$, which regulates the trade-off between reasoning format adherence ($R_f$) and prediction accuracy ($R_a$). 
We conduct experiments with $\lambda$ varying from $0$ to $1$ with a step of $0.2$. 
As illustrated in Figure~\ref{fig:hyperparameter}, the model achieves its global optimum at $\lambda=0.2$ (e.g., ACC-ABSC peaks at $\sim90\%$), surpassing the $\lambda=0$ baseline. This initial gain confirms that a moderate penalty for structural deviations effectively scaffolds the reasoning process, helping the model organize its chain-of-thought. 
However, as $\lambda$ exceeds $0.2$, performance degrades monotonically. This decline suggests that when the reward signal is dominated by formatting constraints ($\lambda > 0.2$), the optimization objective shifts away from semantic correctness, causing the model to prioritize ``looking correct'' (perfect tags) over ``being correct'' (accurate sentiment), thereby validating our choice of $\lambda=0.2$ as the ideal cognitive balance.



\paragraph{Human Evaluation of Sentiment Reasoning.}
Complementing our automatic metrics, we perform a qualitative human evaluation to assess the logical validity of the generated reasoning chains. 
Figure~\ref{fig:human_eval} presents the results from 50 randomly sampled instances, revealing that 94.0\% of the reasoning traces were judged as logically correct and factually consistent with the input text. 
This decisive validation demonstrates that ABSA-R1 successfully learns to articulate high-quality justifications, ensuring that its superior classification performance is grounded in reliable and interpretable decision-making processes.


\section{Conclusions}
In this paper, we introduced \textbf{ABSA-R1}, a reasoning-driven LLM framework that bridges the gap between sentiment prediction and cognitive justification. 
By synergizing reinforcement learning with a \textit{cognition-aligned reward model} and a \textit{performance-driven rejection sampling} strategy, ABSA-R1 learns to internalize the ``reason-before-predict'' paradigm, generating natural language rationales that ground its decisions. 
Empirical results across four benchmarks demonstrate that our approach achieves SOTA performance on both ABSC and AOSTE tasks, proving particularly robust in scenarios requiring complex semantic disambiguation. 
Ultimately, this work advances the frontier of interpretable NLP by endowing models with the capacity to articulate the causal ``why'' behind their predictions, paving the way for more transparent, trustworthy, and human-aligned affective computing systems.

\section*{Acknowledgements}
This research is funded by the National Key Research and Development Program of China Grant (No. 2024YFC3308500), the National Nature Science Foundation of China (No.62477010, No.62577022 and No.62307028), Shanghai Science and Technology Innovation Action Plan (No.24YF2710100), Shanghai Qiji Zhifeng Co., Ltd. (2025-GZL-RGZN-01001), the opening funding of the State Key Laboratory of Disaster Reduction in Civil Engineering (Grant No.SLDRCE24-03), and CIPS-SMP-Zhipu Large Model Fund.

\printbibliography

\end{document}